\documentclass[10pt,twocolumn,letterpaper]{article}
\pdfoutput=1

\usepackage{cvpr}
\usepackage{times}
\usepackage{epsfig}
\usepackage{graphicx}
\usepackage{amsmath}
\usepackage{amssymb}
\usepackage{xspace}
\usepackage{algorithm}
\usepackage{algorithmic}
\usepackage{multirow}

\newcommand{\mypar}[1]{\vspace{-4mm}\paragraph{#1}} 

\newcommand{\hertz}{$62~H\hspace{-.5mm}z$\xspace} 
\newcommand{\maxhertz}{$105~H\hspace{-.5mm}z$\xspace} 
\newcommand{\maxpillars}{12000\xspace} 
\newcommand{\maxpts}{100\xspace} 
\newcommand{\xyres}{0.16\xspace} 
\newcommand{\lidar}{lidar\xspace} 

\usepackage[usenames, dvipsnames]{color} 
\definecolor{purple}{rgb}{0.5, 0.0, 0.5}
\definecolor{orange}{rgb}{1, 0.65, 0}
\definecolor{lightgreen}{rgb}{0.68, 1, 0.18}
\definecolor{darkgreen}{rgb}{0.09, 0.32, 0.24}
\definecolor{darkred}{rgb}{0.6, 0, 0}
\definecolor{brown}{rgb}{0.64, 0.16, 0.16}
\iffalse 
  \newcommand{\holger}[1]{\noindent}
  \newcommand{\oscar}[1]{\noindent}
  \newcommand{\alex}[1]{\noindent}
  \newcommand{\sourabh}[1]{\noindent}
  \newcommand{\done}[1]{\noindent}
  \newcommand{\todo}[1]{\noindent}
\else
  \newcommand{\holger}[1]{\textcolor{blue}{\bf [HC: #1]}}
  \newcommand{\oscar}[1]{\textcolor{orange}{\bf [OB: #1]}}
  \newcommand{\alex}[1]{\textcolor{purple}{\bf [AL: #1]}}
  \newcommand{\sourabh}[1]{\textcolor{brown}{\bf [SV: #1]}}
  \newcommand{\done}[1]{\textcolor{darkgreen}{\bf [Done: #1]}}
  \newcommand{\todo}[1]{\textcolor{red}{\bf [Todo: #1]}}
\fi
\usepackage[normalem]{ulem} 

\newcommand{\figref}[1]{Figure \ref{#1}}
\newcommand{\tableref}[1]{Table \ref{#1}}
\newcommand{\secref}[1]{Section \ref{#1}}
\newcommand{\squeeze}{\vspace{-0.5mm}}

\usepackage[pagebackref=true,breaklinks=true,letterpaper=true,colorlinks,bookmarks=false]{hyperref}

\cvprfinalcopy 

\def\cvprPaperID{6374} 

\ifcvprfinal\fi
\begin{document}

\title{PointPillars: Fast Encoders for Object Detection from Point Clouds}

\author{
Alex H. Lang \and Sourabh Vora \and  Holger Caesar \and  Lubing Zhou \and  Jiong Yang \and  Oscar Beijbom \\
nuTonomy: an APTIV company\\
{\tt\small \{alex, sourabh, holger, lubing, jiong.yang, oscar\}@nutonomy.com}
}


\maketitle

\begin{abstract}
Object detection in point clouds is an important aspect of many robotics applications such as autonomous driving. 
In this paper we consider the problem of encoding a point cloud into a format appropriate for a downstream detection pipeline.
Recent literature suggests two types of encoders; fixed encoders tend to be fast but sacrifice accuracy, while encoders that are learned from data are more accurate, but slower.
In this work we propose PointPillars, a novel encoder which utilizes PointNets to learn a representation of point clouds organized in vertical columns (pillars).
While the encoded features can be used with any standard 2D convolutional detection architecture, we further propose a lean downstream network.
Extensive experimentation shows that PointPillars outperforms previous encoders with respect to both speed and accuracy by a large margin.
Despite only using lidar, our full detection pipeline significantly outperforms the state of the art, even among fusion methods, with respect to both the 3D and bird's eye view KITTI benchmarks.
This detection performance is achieved while running at \hertz: a 2 - 4 fold runtime improvement.
A faster version of our method matches the state of the art at 105 Hz.
These benchmarks suggest that PointPillars is an appropriate encoding for object detection in point clouds.
\end{abstract}

\begin{figure}
\begin{center}
\includegraphics[width = 8.3cm]{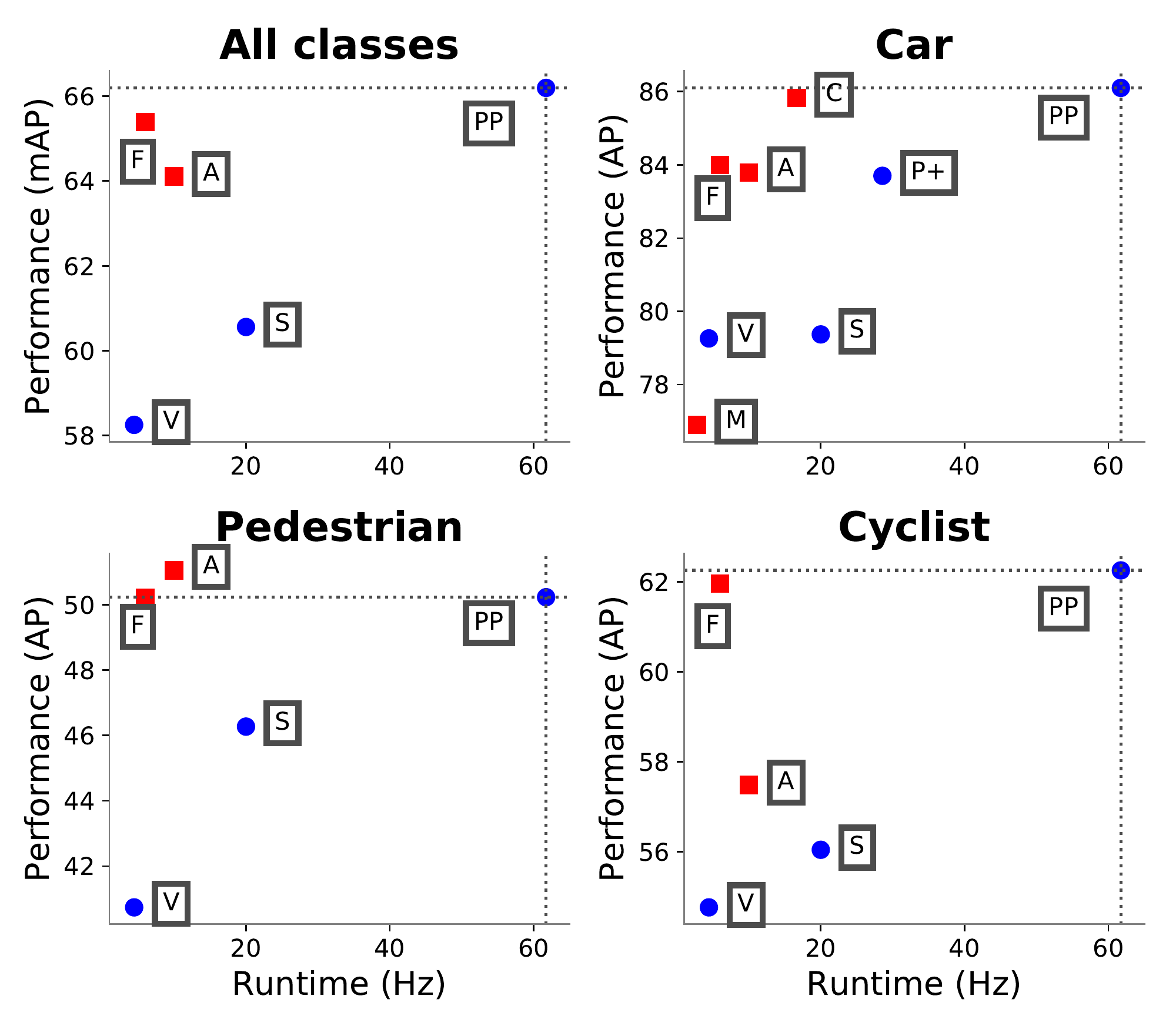}
\end{center}
\squeeze
\caption{Bird's eye view performance vs speed for our proposed PointPillars, \framebox{PP} method on the KITTI~\cite{kitti} test set.
Lidar-only methods drawn as blue circles; \lidar \& vision methods drawn as red squares.
Also drawn are top methods from the KITTI leaderboard: \framebox{M}: MV3D~\cite{mv3d}, \framebox{A} AVOD~\cite{avod}, \framebox{C}: ContFuse~\cite{contfuse}, \framebox{V}: VoxelNet~\cite{voxelnet}, \framebox{F}: Frustum PointNet~\cite{frustum}, \framebox{S}: SECOND~\cite{second}, \framebox{P+} PIXOR++~\cite{hdnet}.
PointPillars outperforms all other \lidar-only methods in terms of both speed and accuracy by a large margin.
It also outperforms all fusion based method except on pedestrians. 
Similar performance is achieved on the 3D metric (\tableref{table:res_3d}). 
 }
\label{fig:speed_accuracy}
\end{figure}

\squeeze
\section{Introduction} \label{sec:intro}
\squeeze

Deploying autonomous vehicles (AVs) in urban environments poses a difficult technological challenge.
Among other tasks, AVs need to detect and track moving objects such as vehicles, pedestrians, and cyclists in realtime.
To achieve this, autonomous vehicles rely on several sensors out of which the \lidar is arguably the most important.
A \lidar uses a laser scanner to measure the distance to the environment, thus generating a sparse point cloud representation.
Traditionally, a \lidar robotics pipeline interprets such point clouds as object detections through a bottom-up pipeline involving background subtraction, followed by spatiotemporal clustering and classification~\cite{classic_lidar,himmelsbach2008lidar}.

Following the tremendous advances in deep learning methods for computer vision,
a large body of literature has investigated to what extent this technology could be applied towards object detection from \lidar point clouds~\cite{voxelnet,hdnet,pixor,avod,mv3d,frustum,contfuse,second,complexyolo,roarnet}.
While there are many similarities between the modalities, there are two key differences:
1) the point cloud is a sparse representation, while an image is dense and
2) the point cloud is 3D, while the image is 2D.
As a result object detection from point clouds does not trivially lend itself to standard image convolutional pipelines.

Some early works focus on either using 3D convolutions~\cite{engelcke2017vote3deep} or a projection of the point cloud into the image \cite{fcl}.
Recent methods tend to view the \lidar point cloud from a bird's eye view~\cite{mv3d, avod, voxelnet, pixor}.
This overhead perspective offers several advantages such as lack of scale ambiguity and the near lack of occlusion.

However, the bird's eye view tends to be extremely sparse which makes direct application of convolutional neural networks impractical and inefficient.
A common workaround to this problem is to partition the ground plane into a regular grid,
for example 10 x 10 cm, and then perform a hand crafted feature encoding method on the points in each grid cell~\cite{mv3d, avod, complexyolo, pixor}.
However, such methods may be sub-optimal since the hard-coded feature extraction method may not generalize to new configurations without significant engineering efforts.
To address these issues, and building on the PointNet design developed by Qi et al.~\cite{pointnet}, VoxelNet~\cite{voxelnet} was one of the first methods to truly do end-to-end learning in this domain.
VoxelNet divides the space into voxels, applies a PointNet to each voxel, followed by a 3D convolutional middle layer to consolidate the vertical axis, after which a 2D convolutional detection architecture is applied.
While the VoxelNet performance is strong, the inference time, at $4.4~H\hspace{-.5mm}z$, is too slow to deploy in real time. Recently SECOND~\cite{second} improved the inference speed of VoxelNet but the 3D convolutions remain a bottleneck.

In this work we propose PointPillars: a method for object detection in 3D that enables end-to-end learning with only 2D convolutional layers.
PointPillars uses a novel encoder that learn features on pillars (vertical columns) of the point cloud to predict 3D oriented boxes for objects.
There are several advantages of this approach.
First, by learning features instead of relying on fixed encoders, PointPillars can leverage the full information represented by the point cloud.
Further, by operating on pillars instead of voxels there is no need to tune the binning of the vertical direction by hand.
Finally, pillars are highly efficient because all key operations can be formulated as 2D convolutions which are extremely  efficient to compute on a GPU.
An additional benefit of learning features is that PointPillars requires no hand-tuning to use different point cloud configurations.
For example, it can easily incorporate multiple lidar scans, or even radar point clouds.

We evaluated our PointPillars network on the public KITTI detection challenges which require detection of cars, pedestrians, and cyclists in either the bird's eye view (BEV) or 3D~\cite{kitti}.
While our PointPillars network is trained using only \lidar point clouds, it dominates the current state of the art including methods that use lidar \emph{and} images, thus establishing new standards for performance on both BEV and 3D detection (\tableref{table:res_bev} and \tableref{table:res_3d}). At the same time PointPillars runs at \hertz, which is orders of magnitude faster than previous art. PointPillars further enables a trade off between speed and accuracy; in one setting we match state of the art performance at over $100~H\hspace{-.5mm}z$ (\figref{fig:speed_accuracy_val}). We have also released code (\href{https://github.com/nutonomy/second.pytorch}{https://github.com/nutonomy/second.pytorch}) that can reproduce our results.

\begin{figure*}
\begin{center}
\includegraphics[width = 1.0\textwidth]{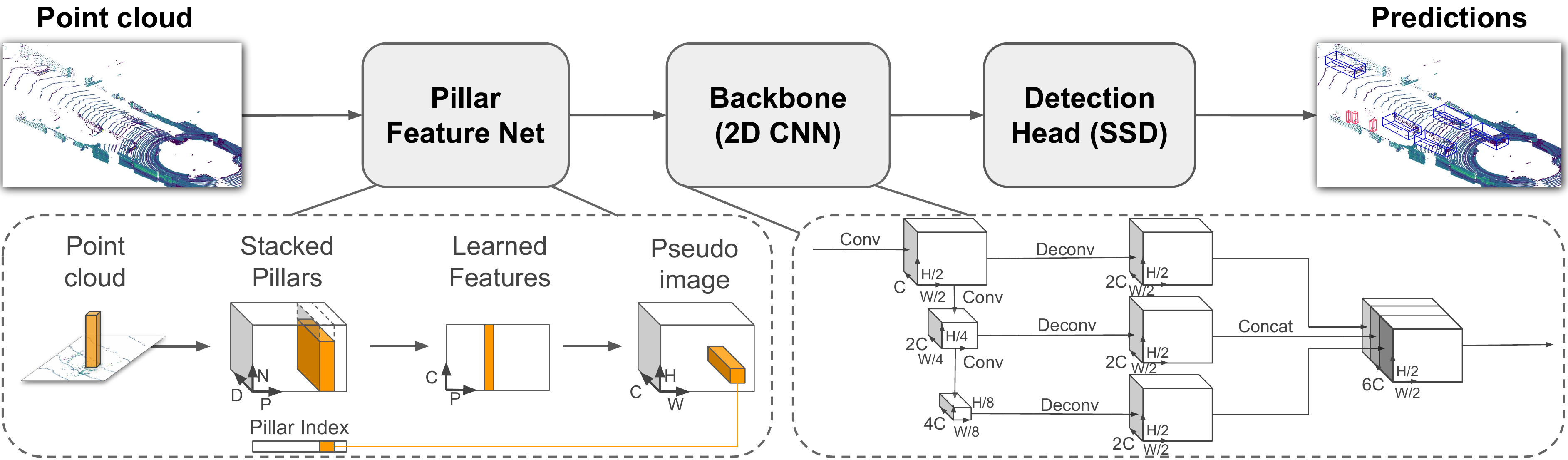}
\end{center}
\squeeze
\caption{Network overview.
The main components of the network are a Pillar Feature Network, Backbone, and SSD Detection Head.
See Section \ref{sec:network} for more details.
The raw point cloud is converted to a stacked pillar tensor and pillar index tensor.
The encoder uses the stacked pillars to learn a set of features that can be scattered back to a 2D pseudo-image for a convolutional neural network.
The features from the backbone are used by the detection head to predict 3D bounding boxes for objects.
Note: here we show the backbone dimensions for the car network.}
\label{fig:network}
\label{fig:pointpillar}
\end{figure*}

\subsection{Related Work}
\squeeze
We start by reviewing recent work in applying convolutional neural networks toward object detection in general, and then focus on methods specific to object detection from \lidar point clouds.

\squeeze
\subsubsection{Object detection using CNNs}
\squeeze
Starting with the seminal work of Girshick et al.~\cite{girshick2014rich} it was established that convolutional neural network~(CNN) architectures are state of the art for detection in images.
The series of papers that followed~\cite{ren2015faster, he2017mask} advocate a two-stage approach to this problem, where in the first stage a region proposal network (RPN) suggests candidate proposals.
Cropped and resized versions of these proposals are then classified by a second stage network.
Two-stage methods dominated the important vision benchmark datasets such as COCO~\cite{coco} over single-stage architectures originally proposed by Liu et al.~\cite{ssd}.
In a single-stage architecture a dense set of anchor boxes is regressed and classified in a single stage into a set of predictions providing a fast and simple architecture.
Recently Lin et al.~\cite{retinanet} convincingly argued that with their proposed focal loss function a single stage method is superior to two-stage methods, both in terms of accuracy \emph{and} runtime. 
In this work, we use a single stage method.

\squeeze
\subsubsection{Object detection in \lidar point clouds}
\squeeze
Object detection in point clouds is an intrinsically three dimensional problem.
As such, it is natural to deploy a 3D convolutional network for detection, which is the paradigm of several early works~\cite{engelcke2017vote3deep,li20173d}.
While providing a straight-forward architecture, these methods are slow; e.g. Engelcke et al.~\cite{engelcke2017vote3deep} require $0.5s$ for inference on a single point cloud.
Most recent methods improve the runtime by projecting the 3D point cloud either onto the ground plane~\cite{avod, mv3d} or the image plane~\cite{fcl}.
In the most common paradigm the point cloud is organized in voxels and the set of voxels in each vertical column is encoded into a fixed-length, hand-crafted, feature encoding to form a pseudo-image which can be processed by a standard image detection architecture.
Some notable works here include MV3D~\cite{mv3d}, AVOD~\cite{avod}, PIXOR~\cite{pixor} and Complex YOLO~\cite{complexyolo} which all use variations on the same fixed encoding paradigm as the first step of their architectures.
The first two methods additionally fuse the \lidar features with image features to create a multi-modal detector.
The fusion step used in MV3D and AVOD forces them to use two-stage detection pipelines, while PIXOR and Complex YOLO use single stage pipelines.

In their seminal work Qi et al.~\cite{pointnet, pointnetplusplus} proposed a simple architecture, PointNet, for learning from unordered point sets, which offered a path to full end-to-end learning.
VoxelNet~\cite{voxelnet} is one of the first methods to deploy PointNets for object detection in \lidar point clouds.
In their method, PointNets are applied to voxels which are then processed by a set of 3D convolutional layers followed by a 2D backbone and a detection head.
This enables end-to-end learning, but like the earlier work that relied on 3D convolutions, VoxelNet is slow, requiring 225ms inference time ($4.4~H\hspace{-.5mm}z$) for a single point cloud.
Another recent method, Frustum PointNet~\cite{frustum}, uses PointNets to segment and classify the point cloud in a frustum generated from projecting a detection on an image into 3D.
Frustum PointNet's achieved high benchmark performance compared to other fusion methods, but its multi-stage design makes end-to-end learning impractical.
Very recently SECOND~\cite{second} offered a series of improvements to VoxelNet resulting in stronger performance and a much improved speed of $20~H\hspace{-.5mm}z$.
However, they were unable to remove the expensive 3D convolutional layers.

\subsection{Contributions}
\squeeze
\begin{itemize}
\setlength\itemsep{1mm}
\item We propose a novel point cloud encoder and network, PointPillars, that operates on the point cloud to enable end-to-end training of a 3D object detection network.
\item We show how all computations on pillars can be posed as dense 2D convolutions which enables inference at \hertz; a factor of 2-4 times faster than other methods.
\item We conduct experiments on the KITTI dataset and demonstrate state of the art results on cars, pedestrians, and cyclists on both BEV and 3D benchmarks.
\item We conduct several ablation studies to examine the key factors that enable a strong detection performance.
\end{itemize}


\begin{figure*}
\begin{center}
\includegraphics[width = 1.0\textwidth]{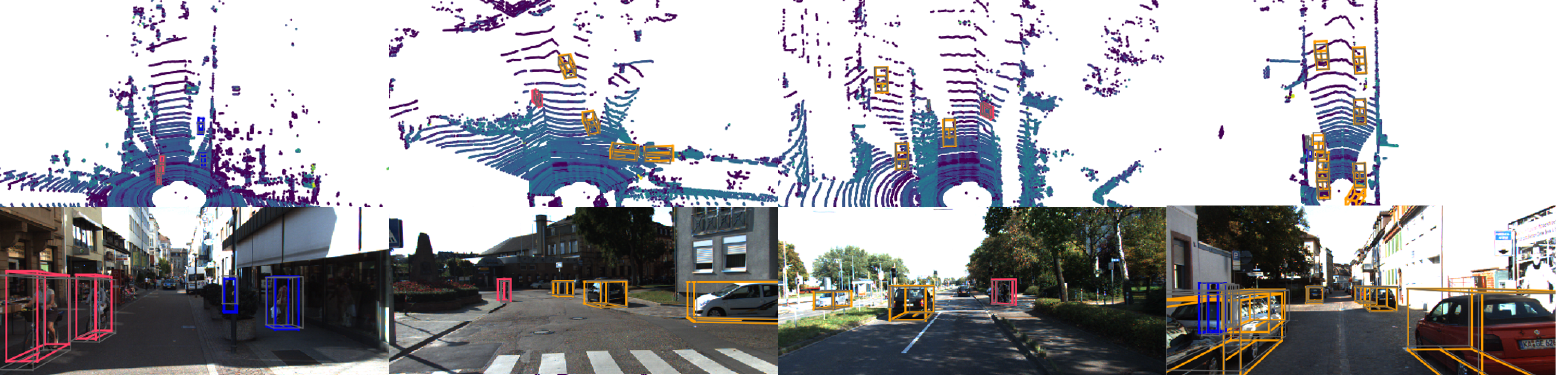}
\end{center}
\caption{Qualitative analysis of KITTI results.
We show a bird's-eye view of the lidar point cloud (top), as well as the 3D bounding boxes projected into the image for clearer visualization.
Note that our method \textit{only} uses lidar.
We show predicted boxes for car (orange), cyclist (red) and pedestrian (blue).
Ground truth boxes are shown in gray.
The orientation of boxes is shown by a line connected the bottom center to the front of the box.
}
\label{fig:kitti_visualize}
\end{figure*}

\begin{figure*}
\begin{center}
\includegraphics[width = 1.0\textwidth]{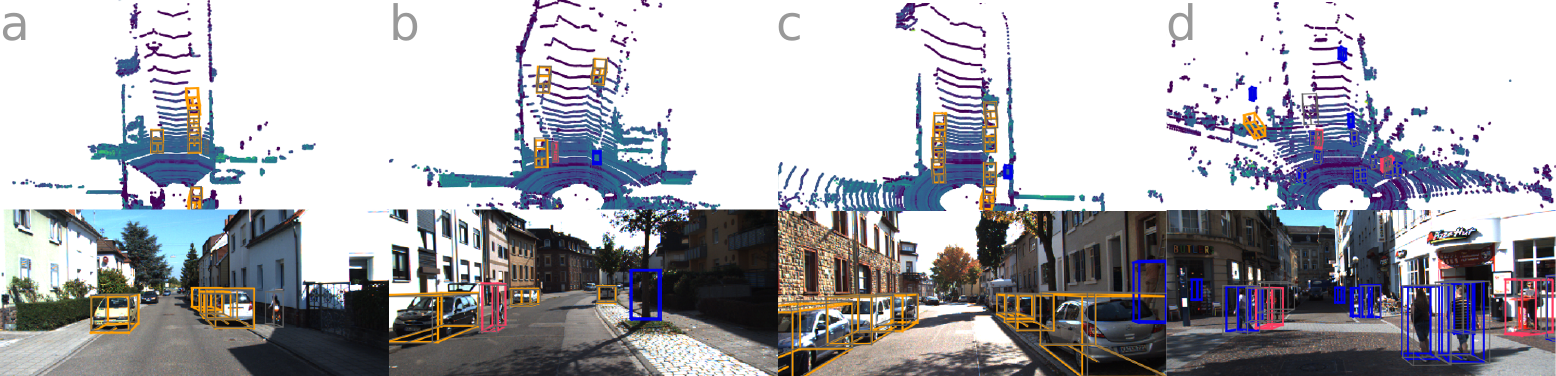}
\end{center}
\caption{Failure cases on KITTI.
Same visualize setup from \figref{fig:kitti_visualize} but focusing on several common failure modes.
}
\label{fig:kitti_failure}
\end{figure*}

\section{PointPillars Network} \label{sec:network}
\squeeze

PointPillars accepts point clouds as input and estimates oriented 3D boxes for cars, pedestrians and cyclists.
It consists of three main stages (\figref{fig:pointpillar}):
(1) A feature encoder network that converts a point cloud to a sparse pseudo-image;
(2) a 2D convolutional backbone to process the pseudo-image into high-level representation; and
(3) a detection head that detects and regresses 3D boxes.

\subsection{Pointcloud to Pseudo-Image}
\label{sec:pointpillars}
\label{sec:create-stacked-pillars}
To apply a 2D convolutional architecture, we first convert the point cloud to a pseudo-image.

We denote by $l$ a point in a point cloud with coordinates $x$, $y$, $z$ and reflectance $r$.
As a first step the point cloud is discretized into an evenly spaced grid in the x-y plane, creating a set of pillars $\mathcal{P}$ with $|\mathcal{P}| = B$.
Note that there is no need for a hyper parameter to control the binning in the z dimension.
The points in each pillar are then augmented with $x_{c}$, $y_{c}$, $z_{c}$, $x_{p}$ and $y_{p}$ where the $c$ subscript denotes distance to the arithmetic mean of all points in the pillar and the $p$ subscript denotes the offset from the pillar $x, y$ center. The augmented lidar point $l$ is now $D=9$ dimensional.

The set of pillars will be mostly empty due to sparsity of the point cloud, and the non-empty pillars will in general have few points in them.
For example, at $0.16^2~m^2$ bins the point cloud from an HDL-64E Velodyne lidar has 6k-9k non-empty pillars in the range typically used in KITTI for $\sim 97\%$ sparsity.
This sparsity is exploited by imposing a limit both on the number of non-empty pillars per sample ($P$) and on the number of points per pillar ($N$) to create a dense tensor of size $(D, P, N)$. 
If a sample or pillar holds too much data to fit in this tensor the data is randomly sampled. 
Conversely, if a sample or pillar has too little data to populate the tensor, zero padding is applied.

Next, we use a simplified version of PointNet where, for each point, a linear layer is applied followed by BatchNorm~\cite{batchnorm} and ReLU~\cite{relu} to generate a $(C, P, N)$ sized tensor. 
This is followed by a max operation over the channels to create an output tensor of size $(C, P)$. 
Note that the linear layer can be formulated as a 1x1 convolution across the tensor resulting in very efficient computation.

\label{sec:stacked-to-pseudo}
Once encoded, the features are scattered back to the original pillar locations to create a pseudo-image of size $(C, H, W)$ where $H$ and $W$ indicate the height and width of the canvas.

\squeeze
\subsection{Backbone}
\squeeze
We use a similar backbone as~\cite{voxelnet} and the structure is shown in Figure \ref{fig:pointpillar}.
The backbone has two sub-networks:
one top-down network that produces features at increasingly small spatial resolution and a second network that performs upsampling and concatenation of the top-down features.
The top-down backbone can be characterized by a series of blocks Block($S$, $L$, $F$).
Each block operates at stride $S$ (measured relative to the original input pseudo-image).
A block has $L$ 3x3 2D conv-layers with $F$ output channels, each followed by BatchNorm and a ReLU.
The first convolution inside the layer has stride $\frac{S}{S_{in}}$ to ensure the block operates on stride $S$ after receiving an input blob of stride $S_{in}$.
All subsequent convolutions in a block have stride 1.

The final features from each top-down block are combined through upsampling and concatenation as follows.
First, the features are upsampled, Up($S_{in}$, $S_{out}$, $F$) from an initial stride $S_{in}$ to a final stride $S_{out}$ (both again measured wrt. original pseudo-image) using a transposed 2D convolution with $F$ final features.
Next, BatchNorm and ReLU is applied to the upsampled features.
The final output features are a concatenation of all features that originated from different strides.

\squeeze
\subsection{Detection Head}
\squeeze
In this paper, we use the Single Shot Detector (SSD)~\cite{ssd} setup to perform 3D object detection.
Similar to SSD, we match the priorboxes to the ground truth using 2D Intersection over Union~(IoU) ~\cite{pascal}.
Bounding box height and elevation were not used for matching; instead given a 2D match, the height and elevation become additional regression targets.


\section{Implementation Details} \label{sec:training}
\squeeze
In this section we describe our network parameters and the loss function that we optimize for.

\subsection{Network}
\squeeze
Instead of pre-training our networks, all weights were initialized randomly using a uniform distribution as in~\cite{heInit}. 

The encoder network has $C=64$ output features.
The car and pedestrian/cyclist backbones are the same except for the stride of the first block ($S=2$ for car, $S=1$ for pedestrian/cyclist).
Both network consists of three blocks, Block1($S$, 4, C), Block2($2S$, 6, 2C), and Block3($4S$, 6, 4C).
Each block is upsampled by the following upsampling steps: Up1($S$, $S$, 2C), Up2($2S$, $S$, 2C) and Up3($4S$, $S$, 2C).
Then the features of Up1, Up2 and Up3 are concatenated together to create 6C features for the detection head.


\subsection{Loss}
\squeeze
We use the same loss functions introduced in SECOND~\cite{second}.
Ground truth boxes and anchors are defined by $(x, y, z, w, l, h, \theta)$.
The localization regression residuals between ground truth and anchors are defined by:
\begin{eqnarray*}
\left.\begin{aligned}
\Delta x &= \frac{x^{gt} - x^a}{d^a}, \Delta y = \frac{y^{gt} - y^a}{d^a}, \Delta z = \frac{z^{gt} - z^a}{h^a} \\
\Delta w &= \log{\frac{w^{gt}}{w^a}}, \Delta l = \log{\frac{l^{gt}}{l^a}}, \Delta h = \log{\frac{h^{gt}}{h^a}} \\
\Delta \theta &= \sin{\left(\theta^{gt} - \theta^a\right)},
\end{aligned}\right.
\end{eqnarray*}
where $x^{gt}$ and $x^a$ are respectively the ground truth and anchor boxes and $d^a = \sqrt{(w^{a})^2+(l^{a})^2}$.
The total localization loss is:
\begin{eqnarray*}
\mathcal{L}_{loc} = \sum_{b \in (x, y, z, w, l, h, \theta)} \text{SmoothL1}\left(\Delta b\right)
\end{eqnarray*}

Since the angle localization loss cannot distinguish flipped boxes, we use a softmax classification loss on the discretized directions~\cite{second}, $\mathcal{L}_{dir}$, which enables the network to learn the heading.

For the object classification loss, we use the focal loss~\cite{retinanet}:

\begin{eqnarray*}
\mathcal{L}_{cls}= -\alpha_a \left(1 - p^a\right) ^ \gamma \log p^a,
\end{eqnarray*}
\squeeze
where $p^a$ is the class probability of an anchor.
We use the original paper settings of $\alpha=0.25$ and $\gamma=2$.
The total loss is therefore: 
\squeeze
\begin{eqnarray*}
\mathcal{L} &= \frac{1}{N_{pos}}\left(\beta_{loc} \mathcal{L}_{loc} + \beta_{cls} \mathcal{L}_{cls}  + \beta_{dir} \mathcal{L}_{dir}\right),
\end{eqnarray*}
\squeeze
where $N_{pos}$ is the number of positive anchors and $\beta_{loc}=2$, $\beta_{cls}=1$, and $\beta_{dir}=0.2$.

To optimize the loss function we use the Adam optimizer with an initial learning rate of $2*10^{-4}$ and decay the learning rate by a factor of $0.8$ every $15$ epochs and train for $160$ epochs.
We use a batch size of 2 for validation set and 4 for our test submission.

\begin{table*}[]
\small
\begin{tabular}{| c | c | c | c || c | c | c || c | c | c || c | c | c |}
\hline
\multirow{2}{*}{Method}				& \multirow{2}{*}{Modality}			& \multirow{2}{10mm}{Speed (Hz)}	& mAP		& \multicolumn{3}{|c||}{Car}		& \multicolumn{3}{|c||}{Pedestrian}		& \multicolumn{3}{|c|}{Cyclist}	\\ \cline{4-13}
					&				& 		& Mod.		& Easy		& Mod.		& Hard		& Easy		& Mod.		& Hard		& Easy		& Mod.		& Hard	\\ \hline 
MV3D~\cite{mv3d}		& Lidar \& Img.	& 2.8		& N/A		& 86.02 		& 76.90 		& 68.49		&  N/A		&  N/A 		&  N/A		&  N/A 		&  N/A 		&  N/A	\\ 
Cont-Fuse~\cite{contfuse}		& Lidar \& Img.	& 16.7			& N/A		& 88.81 		& 85.83 		& 77.33		&  N/A		&  N/A 		&  N/A		&  N/A 		&  N/A 		&  N/A	\\ 
Roarnet~\cite{roarnet}		& Lidar \& Img.	& 10		& N/A		& 88.20 		& 79.41 		& 70.02		&  N/A		&  N/A 		&  N/A		&  N/A 		&  N/A 		&  N/A	\\ 
AVOD-FPN~\cite{avod}	& Lidar \& Img.	& 10		& 64.11		& 88.53		& 83.79 		& 77.90		& \textbf{58.75}	& \textbf{51.05}	& \textbf{47.54}	& 68.09 		& 57.48		& 50.77	\\ 
F-PointNet~\cite{frustum}	&  Lidar \& Img.	& 5.9		& 65.39		& 88.70		& 84.00 		& 75.33		& 58.09		& 50.22		& 47.20		& 75.38		& 61.96		& 54.68	\\ \hline
HDNET~\cite{hdnet}	&  Lidar \& Map	& 20		& N/A		& 89.14		& \textbf{86.57}		& 78.32		& N/A		& N/A		& N/A		& N/A		& N/A		& N/A	\\ \hline
PIXOR++~\cite{hdnet}	& Lidar			& 35			& N/A		& \textbf{89.38} & 83.70 		& 77.97		&  N/A		&  N/A 		&  N/A		&  N/A 		&  N/A 		&  N/A	\\ 
VoxelNet~\cite{voxelnet}	& Lidar			& 4.4		& 58.25		& 89.35	& 79.26 		& 77.39		& 46.13		& 40.74		& 38.11		& 66.70		& 54.76		& 50.55	\\
SECOND~\cite{second}	& Lidar			& 20			& 60.56		& 88.07 		& 79.37 		& 77.95		& 55.10 		& 46.27		& 44.76		& 73.67		& 56.04		& 48.78	\\ \hline
PointPillars			& Lidar			& \textbf{62}	& \textbf{66.19}	& 88.35		& 86.10	& \textbf{79.83}	& 58.66		& 50.23		& 47.19		& \textbf{79.14} & \textbf{62.25} & \textbf{56.00} \\ \hline
\end{tabular}
\caption{Results on the KITTI test BEV detection benchmark.}
\label{table:res_bev}
\end{table*}

\begin{table*}[]
\small
\begin{tabular}{| c | c | c || c || c | c | c || c | c | c || c | c | c |}
\hline
\multirow{2}{*}{Method}				& \multirow{2}{*}{Modality}			& \multirow{2}{10mm}{Speed (Hz)}	& mAP		& \multicolumn{3}{|c||}{Car}		& \multicolumn{3}{|c||}{Pedestrian}		& \multicolumn{3}{|c|}{Cyclist}	\\ \cline{4-13}
					&				& 		& Mod.		& Easy		& Mod.		& Hard		& Easy		& Mod.		& Hard		& Easy		& Mod.		& Hard	\\ \hline 
MV3D~\cite{mv3d}		& Lidar \& Img.	& 2.8		& N/A		& 71.09		& 62.35 		& 55.12		&  N/A		&  N/A 			&  N/A		&  N/A 		&  N/A 		&  N/A	\\ 
Cont-Fuse~\cite{contfuse}		& Lidar \& Img.	& 16.7			& N/A		& 82.54		&  66.22		& 64.04		&  N/A		&  N/A 			&  N/A		&  N/A 		&  N/A 		&  N/A	\\ 
Roarnet~\cite{roarnet}	& Lidar \& Img.	& 10			& N/A		& \textbf{83.71}		&  73.04		& 59.16		&  N/A		&  N/A 			&  N/A		&  N/A 		&  N/A 		&  N/A	\\ 
AVOD-FPN~\cite{avod}	& Lidar \& Img.	& 10		& 55.62		& 81.94 	 	& 71.88	 	& 66.38		& 50.80		& 42.81			& 40.88	& 64.00	 	& 52.18		& 46.61	\\ 
F-PointNet~\cite{frustum}	&  Lidar \& Img.	& 5.9		& 57.35		& 81.20		& 70.39 	 	& 62.19		& 51.21		& \textbf{44.89}		& 40.23		& 71.96	& 56.77 & 50.39	\\ \hline
VoxelNet~\cite{voxelnet}	& Lidar			& 4.4		& 49.05		& 77.47		& 65.11  		& 57.73		& 39.48		& 33.69			& 31.5		& 61.22		& 48.36		& 44.37	\\
SECOND~\cite{second}	& Lidar			& 20			& 56.69		& 83.13 	& 73.66	 	& 66.20		& 51.07		& 42.56			& 37.29		& 70.51		& 53.85 		& 46.90 	\\ \hline
PointPillars			& Lidar			& \textbf{62} 	& \textbf{59.20}	& 79.05		& \textbf{74.99}	& \textbf{68.30}	& \textbf{52.08}	& 43.53			& \textbf{41.49}		& \textbf{75.78} 		& \textbf{59.07}		& \textbf{52.92}	\\ \hline
\end{tabular}
\caption{Results on the KITTI test 3D detection benchmark.}
\label{table:res_3d}
\end{table*}

\begin{table*}[]
\small
\begin{tabular}{| c | c | c || c || c | c | c || c | c | c || c | c | c |}
\hline
\multirow{2}{*}{Method}				& \multirow{2}{*}{Modality}			& \multirow{2}{10mm}{Speed (Hz)}	& mAOS		& \multicolumn{3}{|c||}{Car}		& \multicolumn{3}{|c||}{Pedestrian}		& \multicolumn{3}{|c|}{Cyclist}	\\ \cline{4-13}
					&				& 		& Mod.		& Easy		& Mod.		& Hard		& Easy		& Mod.		& Hard		& Easy		& Mod.		& Hard	\\ \hline 
SubCNN~\cite{xiang2017subcategory}	& Img.	& 0.5		& \textbf{72.71}		&  \textbf{90.61} 	 	& 88.43 	 	& 78.63		&  \textbf{78.33}		& \textbf{66.28}	& \textbf{61.37} 	& 71.39	 & 63.41	& 56.34 	\\ \hline

AVOD-FPN~\cite{avod}	& Lidar \& Img.	& 10		& 63.19		& 89.95  	 	&  87.13	 	& 79.74		& 53.36 		& 44.92			& 43.77 	& 67.61	 	& 57.53 		& 54.16	\\ \hline
SECOND~\cite{second}	& Lidar			& 20			& 54.53		& 87.84  & 81.31	 	& 71.95		& 51.56		& 43.51			& 38.78		& 80.97 		& 57.20 		& 55.14 	\\ \hline
PointPillars			& Lidar			& \textbf{62} 	& 68.86	& 90.19		& \textbf{88.76}	& \textbf{86.38}	& 58.05	& 49.66		& 47.88		& \textbf{82.43} 		& \textbf{68.16}		& \textbf{61.96 }	\\ \hline
\end{tabular}
\caption{Results on the KITTI test average orientation similarity (AOS) detection benchmark. SubCNN is the best performing image only method, while AVOD-FPN, SECOND, and PointPillars are the only 3D object detectors that predict orientation.}\label{table:res_aos}

\end{table*}

\section{Experimental setup}
\squeeze
In this section we present our experimental setup, including dataset, experimental settings and data augmentation.

\subsection{Dataset}
\squeeze
All experiments use the KITTI object detection benchmark dataset~\cite{kitti}, which consists of samples that have both lidar point clouds and images.
We only train on lidar point clouds, but compare with fusion methods that use both lidar and images.
The samples are originally divided into 7481 training and 7518 testing samples.
For experimental studies we split the official training into 3712 training samples and 3769 validation samples~\cite{chen20153d}, while for our test submission we created a minival set of 784 samples from the validation set and trained on the remaining 6733 samples.
The KITTI benchmark requires detections of cars, pedestrians, and cyclists.
Since the ground truth objects were only annotated if they are visible in the image, we follow the standard convention~\cite{mv3d, voxelnet} of only using \lidar points that project into the image.
Following the standard literature practice on KITTI~\cite{avod,voxelnet,second}, we train one network for cars and one network for both pedestrians and cyclists.

\squeeze
\subsection{Settings}
\squeeze

Unless explicitly varied in an experimental study, we use an xy resolution: \xyres m,  max number of pillars ($P$): \maxpillars, and  max number of points per pillar ($N$): \maxpts.

We use the same anchors and matching strategy as~\cite{voxelnet}.
Each class anchor is described by a width, length, height, and z center, and is applied at two orientations: 0 and 90 degrees.
Anchors are matched to ground truth using the 2D IoU with the following rules.
A positive match is either the highest with a ground truth box, or above the positive match threshold, while a negative match is below the negative threshold.
All other anchors are ignored in the loss.

At inference time we apply axis aligned non maximum suppression (NMS) with an overlap threshold of $0.5$ IoU.
This provides similar performance compared to rotational NMS, but is much faster.

\mypar{Car.}
The x, y, z range is [(0, 70.4), (-40, 40), (-3, 1)] meters respectively.
The car anchor has width, length, and height of (1.6, 3.9, 1.5) m with a z center of -1 m.
Matching uses positive and negative thresholds of 0.6 and 0.45.

\mypar{Pedestrian \& Cyclist.}
The x, y, z range of [(0, 48), (-20, 20), (-2.5, 0.5)] meters respectively.
The pedestrian anchor has width, length, and height of (0.6, 0.8, 1.73) meters with a z center of -0.6 meters, while the cyclist anchor has width, length, and height of (0.6, 1.76, 1.73) meters with a z center of -0.6 meters.
Matching uses positive and negative thresholds of 0.5 and 0.35.

\squeeze
\subsection{Data Augmentation}
\squeeze
Data augmentation is critical for good performance on the KITTI benchmark~\cite{second, pixor, mv3d}.

First, following SECOND~\cite{second}, we create a lookup table of the ground truth 3D boxes for all classes and the associated point clouds that falls inside these 3D boxes.
Then for each sample, we randomly select $15, 0, 8$ ground truth samples for cars, pedestrians, and cyclists respectively and place them into the current point cloud.
We found these settings to perform better than the proposed settings~\cite{second}.

Next, all ground truth boxes are individually augmented.
Each box is rotated (uniformly drawn from $[-\pi/20, \pi/20]$) and translated (x, y, and z independently drawn from $\mathcal{N}(0, 0.25)$) to further enrich the training set.

Finally, we perform two sets of global augmentations that are jointly applied to the point cloud and all boxes.
First, we apply random mirroring flip along the x axis~\cite{pixor}, then a global rotation and scaling~\cite{voxelnet, second}.
Finally, we apply a global translation with x, y, z drawn from $\mathcal{N}(0, 0.2)$ to simulate localization noise.

\section{Results} \label{sec:results}
\squeeze

In this section we present results of our PointPillars method and compare to the literature.

\squeeze
\mypar{Quantitative Analysis.}
All detection results are measured using the official KITTI evaluation detection metrics which are: bird's eye view (BEV), 3D, 2D, and average orientation similarity (AOS).
The 2D detection is done in the image plane and average orientation similarity assesses the average orientation (measured in BEV) similarity for 2D detections.
The KITTI dataset is stratified into easy, moderate, and hard difficulties, and the official KITTI leaderboard is ranked by performance on moderate.

As shown in \tableref{table:res_bev} and \tableref{table:res_3d}, PointPillars outperforms all published methods with respect to mean average precision (mAP).
Compared to lidar-only methods, PointPillars achieves better results across all classes and difficulty strata except for the easy car stratum.
It also outperforms fusion based methods on cars and cyclists.

While PointPillars predicts 3D oriented boxes, the BEV and 3D metrics do not take orientation into account.
Orientation is evaluated using AOS~\cite{kitti}, which requires projecting the 3D box into the image, performing 2D detection matching, and then assessing the orientation of these matches.
The performance of PointPillars on AOS significantly exceeds in all strata as compared to the only two 3D detection methods~\cite{avod,second} that predict oriented boxes (\tableref{table:res_aos}).
In general, image only methods perform best on 2D detection since the 3D projection of boxes into the image can result in loose boxes depending on the 3D pose.
Despite this, PointPillars moderate cyclist AOS of 68.16 outperforms the best image based method~\cite{xiang2017subcategory}.

For comparison to other methods on val, we note that our network achieved BEV AP of ($87.98$, $63.55$, $69.71$) and 3D AP of ($77.98$, $57.86$, $66.02$) for the moderate strata of cars, pedestrians, and cyclists respectively.

\squeeze
\mypar{Qualitative Analysis.}
We provide qualitative results in \figref{fig:kitti_visualize} and \ref{fig:kitti_failure}.
While we only train on lidar point clouds, for ease of interpretation we visualize the 3D bounding box predictions from the BEV and image perspective.
\figref{fig:kitti_visualize} shows our detection results, with tight oriented 3D bounding boxes.
The predictions for cars are particularly accurate and common failure modes include false negatives on difficult samples (partially occluded or faraway objects) or false positives on similar classes (vans or trams).
Detecting pedestrians and cyclists is more challenging and leads to some interesting failure modes.
Pedestrians and cyclists are commonly misclassified as each other (see \figref{fig:kitti_failure}a for a standard example and \figref{fig:kitti_failure}d for the combination of pedestrian and table classified as a cyclist).
Additionally, pedestrians are easily confused with narrow vertical features of the environment such as poles or tree trunks (see \figref{fig:kitti_failure}b).
In some cases we correctly detect objects that are missing in the ground truth annotations (see \figref{fig:kitti_failure}c).


\section{Realtime Inference} \label{sec:runtime}
\squeeze
As indicated by our results (\tableref{table:res_bev} and \figref{fig:speed_accuracy_val}), PointPillars represent a significant improvement in terms of inference runtime.
In this section we break down our runtime and consider the different design choices that enabled this speedup.
We focus on the car network, but the pedestrian and bicycle network runs at a similar speed since the smaller range cancels the effect of the backbone operating at lower strides.
All runtimes are measured on a desktop with an Intel i7 CPU and a 1080ti GPU.

The main inference steps are as follows.
First, the point cloud is loaded and filtered based on range and visibility in the images ($1.4~ms$).
Then, the points are organized in pillars and decorated ($2.7~ms$).
Next, the PointPillar tensor is uploaded to the GPU ($2.9~ms$), encoded ($1.3~ms$), scattered to the pseudo-image ($0.1~ms$), and processed by the backbone and detection heads ($7.7~ms$).
Finally NMS is applied on the CPU ($0.1~ms$) for a total runtime of $16.2~ms$.

\mypar{Encoding.}
The key design to enable this runtime is the PointPilar encoding.
For example, at $1.3~ms$ it is 2 orders of magnitude faster than the VoxelNet encoder ($190~ms$)~\cite{voxelnet}.
Recently, SECOND proposed a faster sparse version of the VoxelNet encoder for a total network runtime of $50~ms$.
They did not provide a runtime analysis, but since the rest of their architecture is similar to ours, it suggests that the encoder is still significantly slower; in their open source implementation\footnote{https://github.com/traveller59/second.pytorch/} the encoder requires $48~ms$.

\mypar{Slimmer Design.}
We opt for a single PointNet in our encoder, compared to 2 sequential PointNets suggested by~\cite{voxelnet}.
This reduced our runtime by $2.5~ms$ in our PyTorch runtime.
The number of dimensions of the first block were also lowered $64$ to match the encoder output size, which reduced the runtime by $4.5~ms$.
Finally, we saved another $3.9~ms$ by cutting the output dimensions of the upsampled feature layers by half to $128$.
Neither of these changes affected detection performance.

\mypar{TensorRT.}
While all our experiments were performed in PyTorch~\cite{pytorch}, the final GPU kernels for encoding, backbone and detection head were built using NVIDIA TensorRT, which is a library for optimized GPU inference. 
Switching to TensorRT gave a $45.5\%$ speedup from the PyTorch pipeline which runs at $42.4~H\hspace{-.5mm}z$.

\mypar{The Need for Speed.}
As seen in \figref{fig:speed_accuracy_val}, PointPillars can achieve \maxhertz with limited loss of accuracy.
While it could be argued that such runtime is excessive since a \lidar typically operates at $20~H\hspace{-.5mm}z$, there are two key things to keep in mind.
First, due to an artifact of the KITTI ground truth annotations, only \lidar points which projected into the front image are utilized, which is only $\sim 10\%$ of the entire point cloud. However, an operational AV needs to view the full environment and process the complete point cloud, significantly increasing all aspects of the runtime.
Second, timing measurements in the literature are typically done on a high-power desktop GPU. However, an operational AV may instead use embedded GPUs or embedded compute which may not have the same throughput.


\section{Ablation Studies} \label{sec:exp}
\squeeze
In this section we provide ablation studies and discuss our design choices compared to the recent literature.

\squeeze
\subsection{Spatial Resolution}
\begin{figure}
\begin{center}
\includegraphics[width = 6cm]{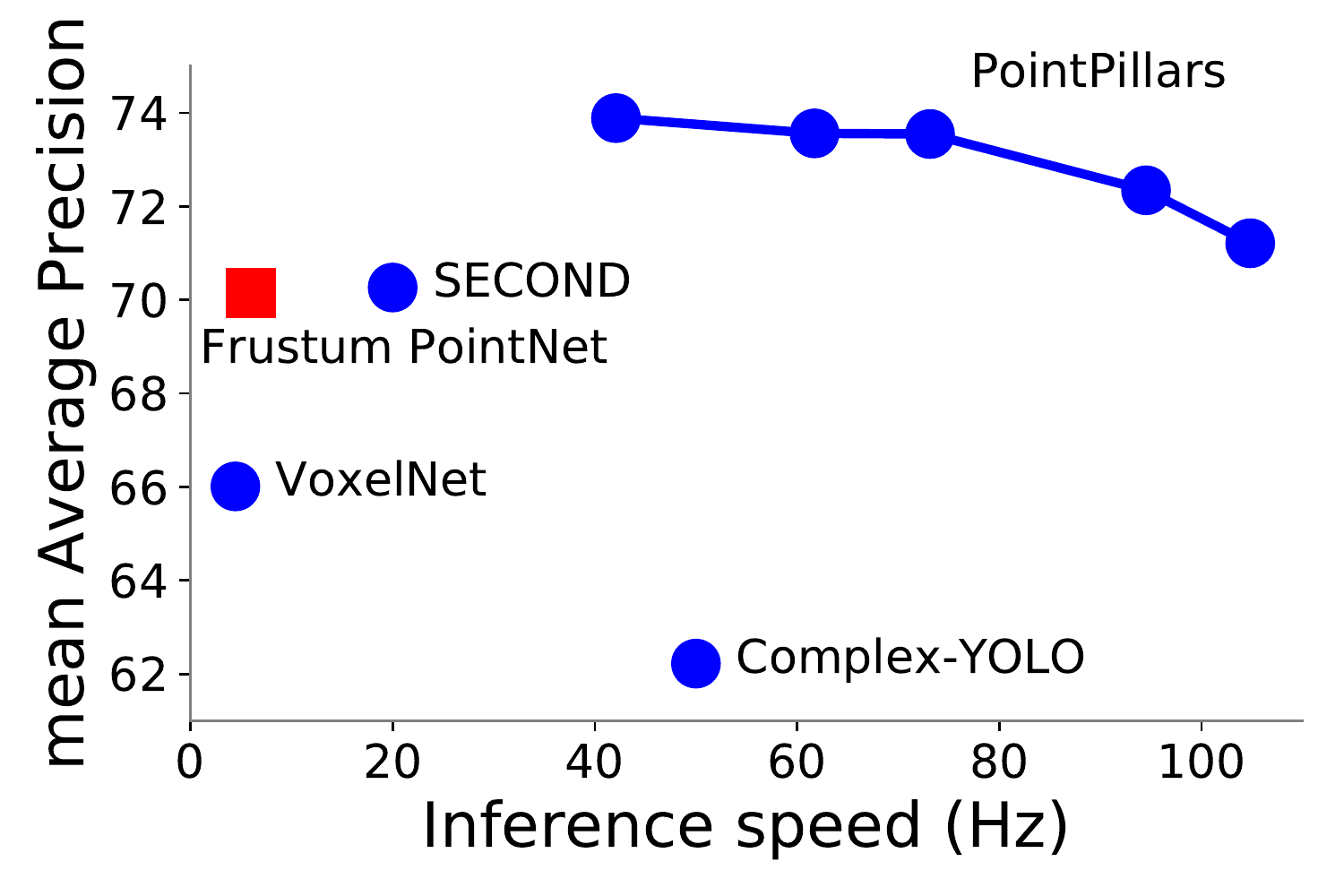}
\end{center}
\vspace{-3mm}
\caption{BEV detection performance (mAP) vs speed (Hz) on the KITTI~\cite{kitti} val set across pedestrians, bicycles and cars.
Blue circles indicate \lidar only methods, red squares indicate methods that use \lidar \& vision.
Different operating points were achieved by using pillar grid sizes in $\{0.12^2, 0.16^2, 0.2^2, 0.24^2, 0.28^2\}$ $m^2$. The number of max-pillars was varied along with the resolution and set to $16000, 12000, 12000, 8000, 8000$ respectively.
}
\label{fig:speed_accuracy_val}
\end{figure}

\squeeze
A trade-off between speed and accuracy can be achieved by varying the size of the spatial binning.
Smaller pillars allow finer localization and lead to more features, while larger pillars are faster due to fewer non-empty pillars (speeding up the encoder) and a smaller pseudo-image (speeding up the CNN backbone).
To quantify this effect we performed a sweep across grid sizes.
From \figref{fig:speed_accuracy_val} it is clear that the larger bin sizes lead to faster networks; at $0.28^2$ we achieve \maxhertz at similar performance to previous methods.
The decrease in performance was mainly due to the pedestrian and cyclist classes, while car performance was stable across the bin sizes.

\squeeze
\subsection{Per Box Data Augmentation}
\squeeze
\label{sec:box_aug}
Both VoxelNet~\cite{voxelnet} and SECOND~\cite{second} recommend extensive per box augmentation.
However, in our experiments, minimal box augmentation worked better.
In particular, the detection performance for pedestrians degraded significantly with more box augmentation.
Our hypothesis is that the introduction of ground truth sampling mitigates the need for extensive per box augmentation.

\squeeze
\subsection{Point Decorations}
\squeeze
\label{sec:decoration}
During the lidar point decoration step, we perform the VoxelNet~\cite{voxelnet} decorations plus two additional decorations:
$x_{p}$ and $y_{p}$ which are the $x$ and $y$ offset from the pillar $x, y$ center.
These extra decorations added 0.5 mAP to final detection performance and provided more reproducible experiments.

\squeeze
\subsection{Encoding}
\squeeze
\label{sec:encoding}
To assess the impact of the proposed PointPillar encoding in isolation, we implemented several encoders in the official codebase of SECOND~\cite{second}.
For details on each encoding, we refer to the original papers.
\begin{table}[]
\small
\begin{tabular}{*{6}{| c} |}
\hline
\textbf{Encoder} 				& \textbf{Type}		& $\mathbf{0.16^2}$ 		&$\mathbf{0.20^2}$  		& $\mathbf{0.24^2}$		& $\mathbf{0.28^2}$		\\ \hline \hline 
MV3D~\cite{mv3d}				& Fixed			& 72.8				& 71.0				& 70.8			& 67.6			\\ \hline
C. Yolo~\cite{complexyolo}		& Fixed			& 72.0		    		& 72.0				& 70.6			& 66.9			\\ \hline
PIXOR~\cite{pixor} 				& Fixed			& 72.9		    		& 71.3				& 69.9			& 65.6			\\ \hline \hline
VoxelNet~\cite{voxelnet}			& Learned			& \textbf{74.4}				& \textbf{74.0}				& \textbf{72.9}			& 71.9 	\\ \hline
PointPillars					& Learned			& 73.7				& 72.6				& \textbf{72.9}			& \textbf{72.0}			\\ \hline
\end{tabular}
\caption{Encoder performance evaluation.
To fairly compare encoders, the same network architecture and training procedure was used and only the encoder and xy resolution were changed between experiments.
Performance is measured as BEV mAP on KITTI val.
Learned encoders clearly beat fixed encoders, especially at larger resolutions.
}
\label{table:encode}
\end{table}

As shown in \tableref{table:encode}, learning the feature encoding is strictly superior to fixed encoders across all resolution.
This is expected as most successful deep learning architectures are trained end-to-end.
Further, the differences increase with larger bin sizes where the lack of expressive power of the fixed encoders are accentuated due to a larger point cloud in each pillar.
Among the learned encoders VoxelNet is marginally stronger than PointPillars.
However, this is not a fair comparison, since the VoxelNet encoder is orders of magnitude slower and has orders of magnitude more parameters.
When the comparison is made for a similar \emph{inference time}, it is clear that PointPillars offers a better operating point (\figref{fig:speed_accuracy_val}).

There are a few curious aspects of \tableref{table:encode}.
First, despite notes in the original papers that their encoder only works on cars, we found that the MV3D~\cite{mv3d} and PIXOR~\cite{pixor} encoders can learn pedestrians and cyclists quite well.
Second, our implementations beat the respective published results by a large margin ($1-10$ mAP).
While this is not an apples to apples comparison since we only used the respective \emph{encoders} and not the full network architectures, the performance difference is noteworthy.
We see several potential reasons.
For VoxelNet and SECOND we suspect the boost in performance comes from improved data augmentation hyperparameters as discussed in \secref{sec:box_aug}.
Among the fixed encoders, roughly half the performance increase can be explained by the introduction of ground truth database sampling~\cite{second}, which we found to boost the mAP by around $3\%$ mAP.
The remaining differences are likely due to a combination of multiple hyperparameters including network design (number of layers, type of layers, whether to use a feature pyramid); anchor box design (or lack thereof~\cite{pixor}); localization loss with respect to 3D and angle; classification loss; optimizer choices (SGD vs Adam, batch size); and more.
However, a more careful study is needed to isolate each cause and effect.

\squeeze
\section{Conclusion} \label{sec:conclusion}
\squeeze

In this paper, we introduce PointPillars, a novel deep network and encoder that can be trained end-to-end on lidar point clouds.
We demonstrate that on the KITTI challenge, PointPillars dominates all existing methods by offering higher detection performance (mAP on both BEV and 3D) at a faster speed.
Our results suggests that PointPillars offers the best architecture so far for 3D object detection from lidar.

{\small
\bibliographystyle{ieee}
\bibliography{references}
}

\end{document}